\documentclass[11pt,a4paper]{article}
\usepackage[hyperref]{style/emnlp2020}
\usepackage{times}
\usepackage{latexsym}

\usepackage{algorithm}
\usepackage[noend]{algpseudocode}
\usepackage{varwidth}
\usepackage{multirow}
\usepackage{tabularx}
\usepackage{subcaption}
\usepackage{amssymb}
\usepackage{amsthm}
\usepackage{bbm}
\usepackage{bm}

\usepackage{microtype}

\aclfinalcopy

\title{Cold-start Active Learning through Self-supervised Language Modeling}
\author{Michelle Yuan\thanks{$^{\star}$Work done while visiting
        National Taiwan University.}\\
    University of Maryland\\
    \texttt{myuan@cs.umd.edu} \\\And
    Hsuan-Tien Lin \\
    National Taiwan University \\
\texttt{htlin@csie.ntu.edu.tw} \\\And
    Jordan Boyd-Graber \\
    University of Maryland\\
\texttt{jbg@umiacs.umd.edu} \\}

\date{}

\newif\ifcomment\commenttrue
\usepackage{amsfonts}
\usepackage{amsmath}
\usepackage{amssymb}
\usepackage{bm}
\usepackage{booktabs}
\usepackage[normalem]{ulem}
\newcommand{\abr}[1]{\textsc{#1}}
\usepackage{graphicx}
\usepackage{multirow}
\usepackage{multicol}
\usepackage[T1]{fontenc}

\newcommand{\fone}{$F_1$}
\newcommand{\bert}{\abr{bert}}

\newcommand{\figfile}[1]{2020_emnlp_alps/figures/#1}
\newcommand{\autofig}[1]{2020_emnlp_alps/auto_fig/#1}
\newcommand{\alps}{\abr{alps}}
\newcommand{\coreset}{\abr{coreset}}
\newcommand{\badge}{\abr{badge}}
\newcommand{\km}{$k$-\abr{means}}
\newcommand{\kmpp}{\km++}
\newcommand{\mlm}{\abr{mlm}}
\newcommand{\al}{\abr{al}}
\newcommand{\norm}[1]{\left\lVert#1\right\rVert}
\newcommand{\scibert}{\abr{sci}\bert}
\newcommand{\hl}[1]{\colorbox{yellow}{#1}}

\newcommand{\given}{\, | \,}
\newcommand{\sst}{\abr{sst-$2$}}
\newcommand{\agnews}{\abr{ag}~\abr{news}}
\newcommand{\pubmed}{\abr{pubmed}}

\newtheorem{theorem}{Proposition}

\begin{document}
\maketitle
\begin{abstract}
Active learning strives to reduce
annotation costs by choosing the most critical examples to label.
Typically, the active learning strategy is contingent on the
classification model.  For instance, uncertainty sampling depends on poorly
calibrated
model confidence
scores.
In the cold-start setting, active learning is impractical
because of model instability and data scarcity.
Fortunately, modern NLP provides an additional source of information:
pre-trained language models.
The pre-training loss can find examples that surprise the model and should be
labeled for efficient fine-tuning.
Therefore, we treat the language modeling loss as a proxy for classification
uncertainty.
 With \bert, we develop a simple strategy based on
 the masked language modeling loss that minimizes labeling costs for text
 classification.
Compared to other baselines, our approach reaches higher accuracy within less sampling
iterations and computation time.

\end{abstract}
\section{Introduction}

Labeling data is a fundamental bottleneck in machine learning, especially for
\abr{nlp}, due to annotation cost and time.  The goal of active learning
(\al{}) is to recognize the most relevant examples and then query labels from an oracle.
For instance, policymakers and physicians
want to quickly fine-tune a text classifier to understand emerging
medical conditions~\citep{voorhees-2020}. Finding labeled data for medical text
is challenging because of privacy issues or shortage in expertise~\citep{dernoncourt-2017}.
Using \al{}, they can query labels for a small subset of the most relevant documents
and immediately train a robust model.

Modern transformer models dominate the leaderboards for several
\abr{nlp} tasks~\citep{devlin-2019,yang-2019}.
Yet the price of adopting transformer-based models is to use
more data.
If these models are not fine-tuned on enough examples,
their accuracy drastically varies across different hyperparameter
configurations~\citep{dodge-2020}.  Moreover, computational resources are a
major drawback as training one model can cost thousands of dollars in cloud
computing and hundreds of pounds in carbon emissions~\citep{strubell-2019}.
These problems motivate further work in \abr{al} to conserve resources.

Another issue is that traditional \al{} algorithms,
like uncertainty sampling~\citep{lewis-1994}, falter on deep models.
These strategies use model confidence scores, but
 neural networks are poorly calibrated~\citep{guo-2017}. High confidence
 scores do not imply high correctness likelihood, so the sampled examples are
 not the most uncertain ones~\citep{zhang-2017}.  Plus, these
strategies sample one document on
each iteration. The single-document sampling requires training the model after
each query and
increases the overall
expense.

These limitations of modern \abr{nlp} models
illustrate a twofold effect: they show a greater need for \al{} \textit{and} make
\al{} more difficult to deploy.
Ideally, \al{} could be most useful during low-resource
situations. In reality, it is impractical to use because the \al{} strategy depends
on warm-starting the model with information about the task~\citep{ash-2019}.
Thus, a fitting
solution to \abr{al} for deep classifiers is a \emph{cold-start} approach, one
that does not rely on classification loss or confidence scores.

To develop a cold-start \abr{al} strategy, we should extract knowledge from pre-trained models like
\bert{}~\citep{devlin-2019}.
The model encodes syntactic properties~\citep{tenney-2019},
acts as a database for
general world
knowledge~\citep{petroni-2019,davison-2019}, and can detect out-of-distribution
examples~\citep{hendrycks-2020}.
Given the knowledge already encoded in pre-trained models, the annotation for
a new task should focus on the information missing from pre-training.
If a sentence contains many words that perplex the language model, then
it is possibly unusual or not well-represented in the
pre-training data.
Thus, the self-supervised objective serves as a surrogate for classification uncertainty.

We develop \alps{} (Active Learning by Processing Surprisal), an \al{} strategy
for \bert-based models.\footnote{\url{https://github.com/forest-snow/alps}}
While many \al{}
methods randomly choose an initial sample, \alps{}
selects the first batch of data using the masked language modeling loss.
As the highest and most extensive peaks
in Europe are found in the Alps,
the \alps{} algorithm finds examples in the data that are both surprising and substantial.
To the best of our knowledge, \alps{} is the first \al{} algorithm that only
relies on a self-supervised loss function. We evaluate our approach on four text
classification datasets spanning across three different domains.
\alps{}
outperforms  \al{} baselines in accuracy and algorithmic efficiency.
The success of \alps{} highlights the importance of self-supervision
for cold-start \al.

\section{Preliminaries}
\label{sec:prelim}
We formally introduce the setup, notation, and terminology that
will be used throughout the paper.

\paragraph{Pre-trained Encoder}
Pre-training uses the language modeling loss
to train encoder parameters for generalized representations.
We call the model input $x=(w_i)_{i=1}^l$ a
``sentence'', which is a sequence of tokens $w$ from a vocabulary $\mathcal{V}$ with
 sequence length $l$.
Given weights $W$, the encoder $h$ maps $x$ to a
$d$-dimensonal hidden
representation $h(x; W)$.
We use \bert~\citep{devlin-2019} as our data
encoder, so $h$ is pre-trained with two tasks: masked language modeling (\mlm{}) and next sentence
prediction.
The embedding $h(x;W)$ is computed as the final
hidden state of the \texttt{[CLS]} token in $x$. We
also refer to $h(x;W)$ as the \bert{} embedding.

\begin{algorithm}[!t]
    \caption{\abr{al} for Sentence Classification}
\label{alg:active}
\begin{algorithmic}[1]
    \Require Inital model $f(x; \theta_0)$
    with pre-trained encoder $h(x; W_0)$, unlabeled data pool
    $\mathcal{U}$, number of queries per iteration $k$, number of
    iterations $T$, sampling
    algorithm $\mathcal{A}$
    \State $\mathcal{D} = \{\}$
    \For {iterations $t=1,\dots,T$}
        \If{$\mathcal{A}$ is cold-start for iteration $t$}
        \State $M_t(x) = f(x; \theta_0)$
        \Else
        \State $M_t(x) = f(x; \theta_{t-1})$
        \EndIf
        \State $\mathcal{Q}_t \gets$~Apply $\mathcal{A}$ on model $M_t(x)$, data
        $\mathcal{U}$
        \State $\mathcal{D}_t \gets$~Label queries $\mathcal{Q}_t$
        \State $\mathcal{D} = \mathcal{D} \cup \mathcal{D}_t$
        \State $\mathcal{U} = \mathcal{U} \setminus \mathcal{D}_t$
        \State $\theta_t \gets$~Fine-tune $f(x; \theta_0)$
        on $\mathcal{D}$
    \EndFor
    \State \Return $f(x; \theta_T)$
\end{algorithmic}
\end{algorithm}

\paragraph{Fine-tuned Model}
We fine-tune \bert{} on the downstream task by training the pre-trained model and the attached sequence
classification head.
Suppose that $f$ represents the model with the classification head, has
parameters $\theta = (W,V)$, and
maps input $x$
to a $C$-dimensional vector with confidence scores for each label.
Specifically, $f(x; \theta)
= \sigma(V \cdot h(x; W))$ where $\sigma$ is a softmax function.

Let~$D$ be the labeled data for our classification task where the labels belong to set $\mathcal{Y} = \left\{1,...,C\right\}$.
During fine-tuning, we take a base classifier~$f$ with weights~$W_0$ from a
pre-trained encoder~$h$ and fine-tune~$f$ on $D$ for new parameters
$\theta_t$.
Then, the predicted classification label is $\hat{y} = \arg\max_{y \in
\mathcal{Y}} f(x; \theta_t)_{y}$.

\paragraph{\al{} for Sentence Classification}
Assume that there is a large unlabeled dataset~$U = \left\{(x_{i}) \right\}_{i=1}^n$ of~$n$ sentences.
The goal of \al{} is to sample a subset~$D \subset U$ efficiently
so that fine-tuning the classifier~$f$ on subset~$D$ improves test accuracy.
On each iteration~$t$, the learner uses strategy~$\mathcal{A}$ to  acquire $k$ sentences
from dataset~$U$ and queries
for their labels (Algorithm~\ref{alg:active}).
Strategy~$\mathcal{A}$ usually depends on an acquisition model
$M_t$~\citep{lowell-2019}.
If the strategy depends on model warm-starting, then the acquisition model~$M_t$ is $f$ with parameters $\theta_{t-1}$ from the
previous iteration.  Otherwise, we assume that $M_t$ is the
pre-trained model with parameters $\theta_0$.
After~$T$ rounds, we acquire labels for~$Tk$
sentences.  We provide more concrete details about \abr{al} simulation in
Section~\ref{sec:experiments}.

\section{The Uncertainty--Diversity Dichotomy}
\label{sec:dichotomy}
This section provides background on prior work in \al{}. First, we discuss two
general \abr{al} strategies: uncertainty sampling and
diversity sampling.  Then, we explain the dichotomy between the two concepts
and introduce \badge{}~\citep{ash-2020}, a \abr{sota} method that attempts to resolve this
issue.  Finally, we focus on the limitations of \badge{} and other \al{}
strategies to give motivation for
our work.

\citet{dasgupta-2011} describes uncertainty and diversity as the ``two
faces of \abr{al}''.
While uncertainty sampling efficiently searches the hypothesis space by finding
difficult examples to label, diversity sampling exploits heterogeneity
in the feature space~\citep{xu-2003,hu-2010,zalan-2011}.
Uncertainty sampling requires
model warm-starting because it depends on model predictions, whereas
diversity sampling can be a cold-start approach.
A successful \al{} strategy should integrate both aspects, but
its exact implementation is an open research question.
For example, a na\"ive idea is to use a fixed combination of strategies to
sample points. Nevertheless, \citet{hsu-2015} experimentally show that this
approach hampers accuracy.
\badge{} optimizes for both uncertainty and diversity by using
confidence scores and clustering.
This strategy
beats uncertainty-based algorithms~\citep{wang-2014}, sampling through bandit
learning~\citep{hsu-2015}, and
\coreset~\citep{sener-2018}, a diversity-based method for convolutional neural
networks.

\subsection{\badge}
\label{ssec:badge}
The goal of \badge{} is to sample a diverse and uncertain batch of points for
training neural networks.
The algorithm transforms data into representations that encode
model confidence and then clusters these transformed points.
First, an unlabeled point~$x$ passes through the trained model to
obtain its predicted label $\hat{y}$. Next, a \textbf{gradient embedding} $g_x$ is
computed for $x$ such that it embodies the gradient of the cross-entropy loss
on
$(f(x; \theta), \hat{y})$
with respect to the parameters of the model's last layer.
The gradient embedding is
\begin{equation}
\label{eq:badge}
(g_x)_i =  (f(x; \theta)_i- \mathbbm{1} (\hat{y} = i))h(x; W).
\end{equation}
The $i$-th block of~$g_x$ is the hidden representation~$h(x; W)$ scaled by
the difference between model confidence score $f(x; \theta)_i$ and an
indicator function
$\mathbbm{1}$ that indicates whether the predictive label~$\hat{y}$ is label~$i$.
Finally,
\badge{} chooses a batch to sample by applying \kmpp{}~\citep{arthur-2006} on
the gradient embeddings.
These embeddings consist of model confidence scores and hidden
representations, so they encode information about both uncertainty and
the data distribution.
By applying \kmpp{} on the gradient embeddings, the
 chosen examples differ in feature representation and predictive uncertainty.

\subsection{Limitations}
\badge{} combines uncertainty and diversity sampling to profit from advantages
of both methods but also brings the downsides of both:
reliance on warm-starting and computational inefficiency.

\subsubsection{Model Uncertainty and Inference}
\label{ssec:uncertainty}

\citet{dodge-2020} observe that training is highly unstable when fine-tuning
pre-trained language models on small datasets.
Accuracy significantly varies across different random initializations.
The model has not fine-tuned on
enough examples, so model confidence is an
unreliable measure for uncertainty.
While \badge{} improves over
uncertainty-based methods, it still relies on confidence scores $f(x; \theta)_i$ when
computing the gradient embeddings (Equation~\ref{eq:badge}).
Also, it uses
labels inferred by the model to compensate for lack of supervision
in \al{}, but this inference is inaccurate for ill-trained models.
Thus, warm-start methods may suffer from problems with model uncertainty or
inference.

\subsubsection{Algorithmic Efficiency}
Many diversity-based methods involve distance comparison between embedding
representations, but this computation can be expensive, especially in
high-dimensional space.
For instance, \coreset{} is a
farthest-first traversal in the embedding space where it chooses the farthest point
from the set of points already chosen on each iteration~\citep{sener-2018}.
The embeddings may appropriately represent the data, but issues, like the ``curse of
dimensionality''~\citep{beyer-1999} and the ``hubness
problem''~\citep{tomasev-2013}, persist.  As the dimensionality increase, the
distance between any two points converges to the same value.
Moreover, the gradient embeddings in \badge{} have dimensionality of $Cd$ for a
$C$-way classification task with data dimensionality of $d$
(Equation~\ref{eq:badge}).
These issues make distance comparison between gradient embeddings less meaningful and
raises costs to compute those distances.

\begin{figure}[t]
\centering
\includegraphics[width=\linewidth]{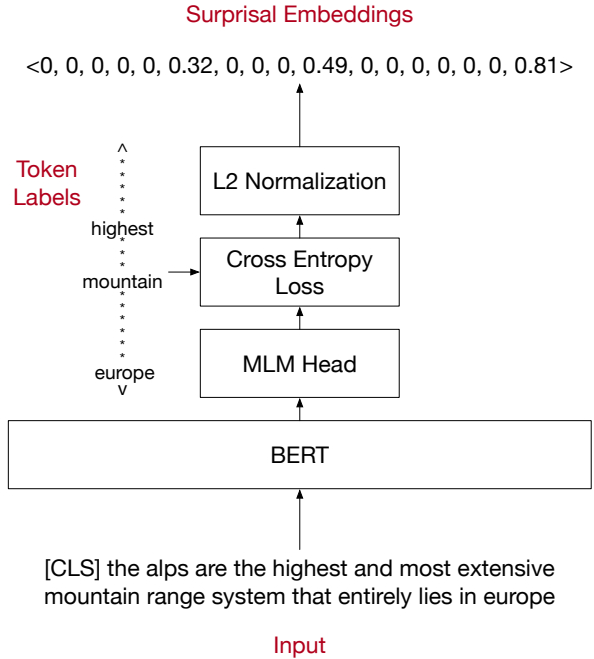}
\caption{To form surprisal embedding $s_x$ for sentence $x$, we pass in unmasked $x$
    through the \bert~\mlm{} head and compute cross-entropy loss for a random
15\% subsample of tokens
against the target labels.  The unsampled tokens have entries of zero in $s_x$.
\alps{} clusters these surprisal embeddings to sample sentences for \al{}.}
\label{fig:mlm}
\end{figure}

\section{A Self-supervised Active Learner}
Cold-start \al{} is challenging because of the shortage in labeled data.
Prior work, like \badge{}, often depend on model uncertainty or inference,
but these measures can be unreliable if the model has not trained on enough
data (Section~\ref{ssec:uncertainty}).
To overcome the lack of supervision, what if we apply
self-supervision to \al{}?  For \abr{nlp}, the language modeling task is
self-supervised because the label for each token is the token itself.  If the task
has immensely improved transfer learning, then it may reduce
generalization error in \al{} too.

For our approach, we adopt the uncertainty-diversity \badge{} framework for
clustering embeddings that encode information about uncertainty.
However, rather than relying on the classification loss gradient, we use the
\mlm{} loss to bootstrap uncertainty estimates.  Thus, we
combine uncertainty and diversity sampling for cold-start \al{}.

\subsection{Masked Language Modeling}

To pre-train \bert{} with \mlm{}, input tokens
are randomly masked, and the model needs to predict the token labels of
the masked tokens.  \bert{} is bidirectional, so it uses context from the left
and right of the masked token to make predictions.
\bert{} also uses next sentence prediction for pre-training, but this
task shows minimal effect for fine-tuning~\citep{liu-2019}.  So, we
focus on applying \mlm{} to \al{}.  The \mlm{} head can
capture syntactic phenomena~\citep{goldberg-2019} and performs well on psycholinguistic
tests~\citep{ettinger-2020}.

\begin{algorithm}[!t]
\caption{Single iteration of \alps}
\label{alg:alps}
\begin{algorithmic}[1]
    \Require Pre-trained encoder $h(x; W_0)$, unlabeled data pool
    $\mathcal{U}$, number of queries $k$
    \For {sentences $x \in \mathcal{U}$}
        \State Compute $s_x$ with \mlm~head of $h(x; W_0)$
    \EndFor
    \State $\mathcal{M} = \{s_x \mid x \in \mathcal{U}\}$
    \State $\mathcal{C} \gets$ \km~cluster centers of $\mathcal{M}$
    \State $\mathcal{Q} = \{\arg\min_{x \in \mathcal{U}} \norm{c-s_x}  | c \in \mathcal{C}  \}$
    \State \Return $\mathcal{Q}$
\end{algorithmic}
\end{algorithm}

\subsection{\alps}

\paragraph{Surprisal Embeddings}
Inspired by how \badge{} forms gradient embeddings from the
classification loss, we create \textbf{surprisal embeddings} from language
modeling.  For sentence $x$, we compute surprisal embedding $s_x$ by evaluating
$x$ with the \mlm{} objective.
To evaluate \mlm{} loss, \bert{} randomly masks 15\% of the tokens in $x$ and
computes cross-entropy loss for the
masked tokens against their true token labels.
When computing surprisal embeddings, we
make one crucial change: \textit{none of the tokens are masked when the input is
passed into \bert}.  However, we still randomly choose 15\% of the tokens in the
input to evaluate with cross-entropy against their target token labels.  The
unchosen tokens are assigned a loss of zero as they are not evaluated (Figure~\ref{fig:mlm}).

These decisions for not masking input (Appendix~\ref{ssec:mask}) and
evaluating only 15\% of tokens (Appendix~\ref{ssec:sample}) are made because of
experiments on the validation set.
Proposition~\ref{theorem:surprisal} provides insight on the information
encoded in surprisal
embeddings.
Finally, the surprisal embedding is $l_2$-normalized as
normalization improves
clustering~\citep{aytekin-2018}.
If the input sentences have a fixed length of $l$, then the surprisal embeddings
have dimensionality of $l$.  The length $l$ is usually less than the hidden
size of \bert{} embeddings.

\begin{theorem}
\label{theorem:surprisal}
For an unnormalized surprisal embedding $s_x$,
each nonzero entry $(s_x)_i$ estimates $I(w_i)$, the surprisal of its
corresponding token within the context of sentence $x$.
\end{theorem}

\begin{proof}
Extending notation from Section~\ref{sec:prelim}, assume that $m$ is the \mlm{} head, with parameters $\phi = (W,
Z)$, which maps
input $x$ to a $l \times |\mathcal{V}|$ matrix $m(x;\phi)$. The $i$th row
$m(x;\phi)_i$
contains prediction scores for $w_i$, the $i$th token in $x$.  Suppose that
$w_i$ is the $j$th token in vocabulary $\mathcal{V}$.
Then, $m(x;\phi)_{i,j}$ is the
likelihood of predicting $w_i$ correctly.

Now, assume that context is the entire input $x$ and define the language model
probability $p_{m}$ as,
\begin{equation}
    p_{m}(w_i \given{} x) = m(x; \phi)_{i,j}.
\label{eq:estimate}
\end{equation}
\citet{salazar-2020} have a similar definition as Equation~\ref{eq:estimate}
but instead have defined it in terms of the masked input.  We argue that
their definition can be extended to the unmasked input $x$. During
\bert{} pre-training, the \mlm{} objective is evaluated on the
\texttt{[MASK]} token for 80\% of the time, random token for 10\% of the
time, and the original token for 10\% of the time. This helps maintain
consistency across pre-training and fine-tuning because \texttt{[MASK]} never appears in
fine-tuning~\citep{devlin-2019}.  Thus, we assume that $m$ estimates occurrence of tokens within a maskless
context as well.

Next, the information-theoretic surprisal~\citep{shannon-1948} is defined as $I(w) = -\log p
(w \given{} c)$, the negative log likelihood of word $w$ given context $c$.
If $w_i$ is sampled and evaluated, then the $i$th entry
of the unnormalized surprisal embedding is,
\begin{align*}
    (s_x)_i &= - \log m(x; \phi)_{i,j} = - \log p_m(w_i \given x) \\
            &= I(w_i).
\end{align*}
\end{proof}

Proposition~\ref{theorem:surprisal} shows that the surprisal embeddings comprise
of estimates for token-context surprisal. Intuitively, these values can help
with \al{} because they highlight the information missing from the pre-trained
model.
For instance, consider the sentences: ``this is my favorite television show''
and ``they feel ambivalent about catholic psychedelic synth folk music''.
Tokens from the latter have higher surprisal than those from the former.  If this is a
sentiment classification task, the second sentence is more confusing for
the classifier to learn.  The surprisal embeddings indicate
sentences challenging for the pre-trained model to understand and
difficult for the fine-tuned model to label.

The most surprising sentences contain many rare tokens.
If we only train our model on the most surprising sentences, then it may
not generalize well across different examples.
Plus, we may sample several atypical
sentences that are similar to each other, which is often an issue for
uncertainty-based methods~\citep{kirsch-2019}.  Therefore, we incorporate clustering in \alps{} to
maintain diversity.

\paragraph{\km~Clustering}
After computing surprisal embeddings for each sentence in the unlabeled pool, we
use \km{} to cluster the surprisal embeddings.  Then, for each
cluster center, we select the
sentence that has the nearest surprisal embedding to it.  The final set of sentences
are the queries to be labeled by an oracle (Algorithm~\ref{alg:alps}).  Although
\badge{} uses \kmpp{} to cluster, experiments show that \km{} works better for
surprisal embeddings (Appendix~\ref{ssec:km}).

\begin{table*}[t]
    \centering
    \begin{tabular}{llrrrr}
    \toprule
    Dataset & Domain & Train & Dev & Test & \# Labels \\
    \midrule
     \agnews & News articles & 110{,}000 & 10{,}000 & 7{,}600 & 4\\
    \abr{imdb} & Sentiment reviews & 17{,}500 & 7{,}500 & 25{,}000 & 2\\
    \pubmed{} 20k \abr{rct} & Medical abstracts & 180{,}040 & 30{,}212 & 30{,}135 & 5\\
    \sst & Sentiment reviews & 60{,}615 & 6{,}736 & 873 & 2 \\
    \bottomrule
    \end{tabular}
    \caption{Sentence classification datasets used in experiments.}
    \label{tab:data}
\end{table*}

\begin{figure*}[t]
\centering
\includegraphics[width=\textwidth]{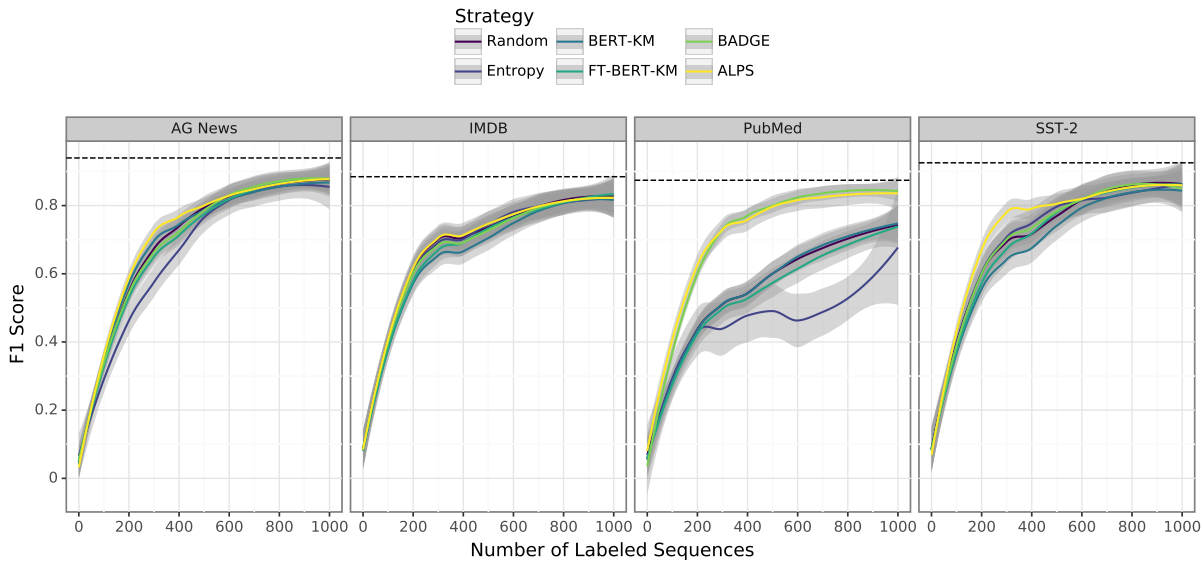}
\caption{Test accuracy of simulated \al{} over ten iterations with 100 sentences
queried per iteration. The dashed line is the test accuracy when the model is
fine-tuned on the entire dataset. Overall, models trained with data sampled from
\alps{} have the highest test accuracy, especially for the earlier iterations.}
\label{fig:seqcls}
\end{figure*}

\section{Active Sentence Classification}
\label{sec:experiments}

We evaluate \alps{} on sentence classification for three different domains:
sentiment reviews, news articles, and medical abstracts (Table~\ref{tab:data}).  To simulate \al{},
we sample a batch of 100 sentences from the training dataset, query labels for this
batch, and then move the
batch from the unlabeled pool to the labeled dataset (Algorithm~\ref{alg:active}).
The initial encoder $h(x; \theta_0)$,
is an already pre-trained, \bert-based model (Section~\ref{ssec:setup}).
In a given
iteration, we fine-tune the
base classifier $f(x; \theta_0)$ on the
labeled dataset and evaluate the fine-tuned model with classification micro-\fone{} score
on the test set.
We do not fine-tune the model $f(x;\theta_{t-1})$ from the previous
iteration to avoid
issues with warm-starting~\citep{ash-2019}.
We repeat
for ten
iterations, collecting a total of 1{,}000 sentences.

\subsection{Baselines}
\label{sec:baselines}
We compare \alps{} against warm-start methods (Entropy, \badge{}, \abr{ft}-\bert{}-\abr{km}) and
cold-start methods (Random, \bert{}-\abr{km}).  For \abr{ft}-\bert{}-\abr{km},
we use \bert{}-\abr{km} to sample data in the first iteration.  For
other warm-start methods, data is randomly sampled in the first iteration.

\paragraph{Entropy} Sample $k$ sentences with highest predictive
entropy measured by $\sum_{i=1}^{C} (f(x; \theta)_i) \ln (f(x; \theta
)_i)^{-1}$~\citep{lewis-1994,wang-2014}.

\paragraph{\badge} Sample $k$ sentences based on diversity in loss gradient
(Section~\ref{ssec:badge}).

\paragraph{\bert{}-{\abr{km}}} Cluster pre-trained, $l_2$-normalized \bert{} embeddings with \km{} and sample the
nearest neighbors of the $k$ cluster centers.  The algorithm is the same as
\alps{} except that \bert{} embeddings are used.

\paragraph{\abr{ft}-\bert{}-\abr{km}}  This is the same algorithm as
\bert{}-\abr{km} except the \bert{} embeddings $h(x; W_{t-1})$ from the
previously fine-tuned model are used.

\subsection{Setup}
\label{ssec:setup}

For each sampling algorithm and dataset, we run the \al{} simulation five
times with different random seeds.  We set the maximum sequence length to 128.  We fine-tune on a
batch size of thirty-two for three epochs.
We use AdamW~\citep{loshchilov-2019} with learning rate of 2e-5,
$\beta_1 = 0.9$, $ \beta_2 = 0.999$, and  a
linear decay of learning rate.

For \abr{imdb}~\citep{maas-2011}, \abr{sst-$2$}~\citep{socher-2013}, and \agnews~\citep{zhang-2015}, the data encoder is the uncased \bert{}-Base model with 110M
parameters.\footnote{\url{https://huggingface.co/transformers/}}
For \pubmed~\citep{dernoncourt-2017}, the data encoder is \scibert, a \bert{} model pre-trained on scientific
texts~\citep{beltagy-2019}.  All experiments are run on GeForce GTX 1080 GPU and
2.6 GHz AMD Opteron 4180 CPU processor; runtimes in Table~\ref{tab:time}.

\subsection{Results}

The model fine-tuned with data sampled by \alps{} has
higher test accuracy than the baselines (Figure~\ref{fig:seqcls}).
For \agnews{}, \abr{imdb}, and \sst{},
this is true in earlier iterations.  We often see the
most gains in the beginning for
crowdsourcing~\citep{felt-2015}.  Interestingly,
clustering the fine-tuned \bert{} embeddings is not always better than
clustering the pre-trained \bert{} embeddings for \abr{al}.
The fine-tuned \bert{} embeddings may require training on more data for more
informative
representations.

For \pubmed, test accuracy greatly varies between the strategies.
The dataset belongs to a specialized domain and is class-imbalanced, so na\"ive
methods show poor accuracy.  Entropy
sampling has the lowest accuracy because the classification entropy is
uninformative in early iterations.
The models
fine-tuned on data sampled by \alps{} and \badge{} have about the same accuracy.
Both methods strive to optimize for uncertainty and diversity, which
alleviates problems with class imbalance.

Our experiments cover the first ten iterations because we focus on the
cold-start setting.  As sampling iterations increase, test accuracy across the
different methods converges.  Both \alps{} and \badge{} already approach the
model trained on the full training dataset across all tasks
(Figure~\ref{fig:seqcls}).  Once the cold-start issue subsides,
uncertainty-based methods can be employed to further query the most
confusing examples for the model to learn.

\begin{table}[t]
    \centering
    \begin{tabular}{lrr}
    \toprule
    & \agnews & \pubmed \\
    \midrule
    Random & <1 & <1\\
    Entropy & 7 & 10 \\
    \alps & 14 & 24 \\
    \badge & 23 & 70\\
    \bert-\abr{km} & 28 & 58 \\
    \abr{ft}-\bert-\abr{km} & 33 & 79 \\
    \bottomrule
    \end{tabular}
    \caption{Average runtime (minutes) per sampling iteration during \al{} simulation for
    large datasets.
    \badge{}, \abr{ft}-\bert{}-\abr{km}, and \bert{}-\abr{km} take much longer
    to run.
    }
    \label{tab:time}
\end{table}

\begin{figure}[t]
    \centering
    \includegraphics[width=\linewidth]{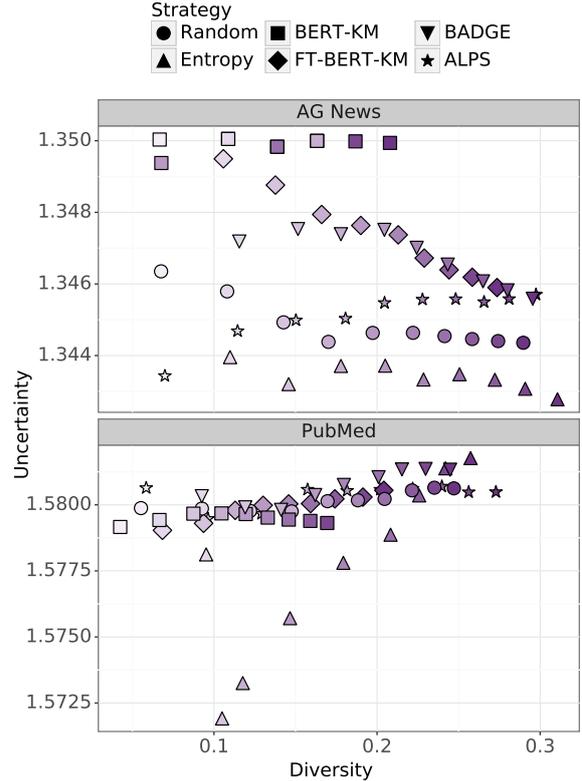}
    \caption{Plot of diversity against uncertainty estimates from \al{}
        simulations for \agnews{} and \pubmed.
        Each point represents a sampled batch
    of sentences from the \al{} experiments.
The shape indicates the strategy used to sample
the sentences.  The color indicates the sample iteration.  The lightest color
corresponds to the first iteration and the darkest color
represents the tenth iteration.  While uncertainty estimates are similar across
different batches, \alps{} shows a consistent increase in diversity without drops in
uncertainty.}
    \label{fig:analysis}
\end{figure}
\begin{figure*}[t]
\centering
    \begin{subfigure}{0.47\linewidth}
	\includegraphics[width=\linewidth]{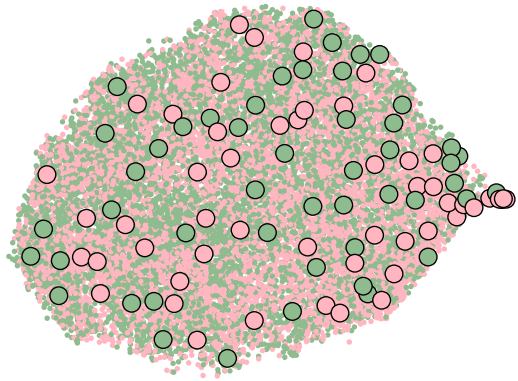}
    \caption{\bert~embeddings with \km~centers}
    \label{fig:embedkm}
    \end{subfigure}
    \begin{subfigure}{0.47\linewidth}
	\includegraphics[width=\linewidth]{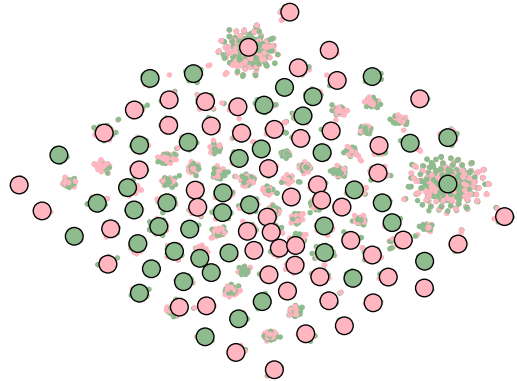}
    \caption{Surprisal embeddings with \km~centers}
    \label{fig:mlmkm}
    \end{subfigure}
    \caption{\abr{t-sne} plots of \bert~embeddings
        and surprisal embeddings for each sequence in the \abr{imdb} training
        dataset.  The enlarged points are the centers determined by \km{} (left)
        and \kmpp{} (right). The points are colored according to their classification labels.
In both sets of embeddings, we cannot clearly separate the points from their
labels, but the distinction between clusters in surprisal embeddings seems more obvious.
}
\label{fig:embed}
\end{figure*}

\section{Analyzing \alps}
\label{sec:analysis}

\paragraph{Sampling Efficiency}
Given that the gradient embeddings are computed, \badge{} has a time complexity of
$\mathcal{O}(Cknd)$ for a $C$-way classification task, $k$ queries, $n$ points in the
unlabeled pool, and $d$-dimensional \bert~embeddings.  Given that the surprisal
embeddings are computed, \alps{} has a time
complexity of $\mathcal{O}(tknl)$ where $t$ is the fixed number of iterations
for \km{} and $l$ is the maximum sequence length.  In our experiments, $k = 100$,
$d=768$, $t=10$, and $l=128$.  In practice, $t$ will not change much, but~$n$
and~$C$
could be much higher.
For large dataset \pubmed, the average runtime per iteration is 24 minutes for
\alps{} and 70 minutes for \badge{} (Table~\ref{tab:time}).  So, \alps{} can match \badge's accuracy more quickly.

\paragraph{Diversity and Uncertainty}

We estimate diversity and uncertainty for data sampled across different
strategies.
For
diversity, we look at the overlap between tokens in the
sampled sentences and tokens from the rest of the data pool.
A diverse
batch of sentences should share many of the same tokens with the data
pool.
In other words, the sampled sentences can represent the data pool because of
the substantial
overlap between their tokens.
In our simulations,
the entire data pool is the training dataset (Section~\ref{sec:experiments}).
So, we compute the Jaccard similarity between $\mathcal{V}_D$, set of
tokens from the sampled sentences $\mathcal{D}$, and $\mathcal{V}_{D'}$, set of
tokens from the unsampled sentences $\mathcal{U} \setminus \mathcal{D}$,
\begin{equation}
    G_d(\mathcal{D}) = J(\mathcal{V}_D, \mathcal{V}_{D'}) =
\frac{|\mathcal{V}_D \cap \mathcal{V}_{D'}|}{|\mathcal{V}_D \cup
\mathcal{V}_{D'}|}.
\end{equation}
If $G_d$ is high, this indicates high diversity because the sampled
and unsampled sentences have many tokens in common.
If $G_d$ is low, this indicates poor diversity and representation.

To measure uncertainty, we use $f(x,\theta_{*})$, the classifier
trained on the full training dataset.   In our experiments, classifier
$f(x,\theta_{*})$ has high accuracy (Figure~\ref{fig:seqcls}) and inference is stable
after training on many examples.  Thus, we can use the logits from the
classifier to understand its uncertainty toward a particular sentence. First, we
compute predictive entropy of sentence $x$ when evaluated by model $f(x,\theta_{*})$.  Then, we
take the average of predictive entropy over all sentences in a sampled batch
$\mathcal{D}$.
We use the average predictive entropy to esimate uncertainty of the sampled
sentences,
\begin{equation}
    G_u(\mathcal{D}) = \frac{1}{|\mathcal{D}|} \sum_{x \in \mathcal{D}} \sum_{i=1}^{C}
    (f(x; \theta_{*})_i) \ln (f(x; \theta_{*})_i)^{-1}.
\end{equation}
We compute $G_d$ and $G_u$ for batches sampled in the \al{}
experiments of \agnews{} and \pubmed.  Diversity is plotted against uncertainty for
batches sampled across different iterations and \al{} strategies
(Figure~\ref{fig:analysis}).  For \agnews{}, $G_d$ and $G_u$ are
relatively low for \alps{} in the first iteration.  As iterations
increase, samples from \alps{} increase in diversity and decrease minimally in
uncertainty.
Samples from other methods have a larger drop in uncertainty as iterations
increase.
For \pubmed, \alps{} again increases in sample diversity without
drops in uncertainty.  In the last iteration, \alps{} has the highest diversity
among all the algorithms.

\paragraph{Surprisal Clusters}

Prior work use \km{} to cluster feature representations as a cold-start
\al{} approach~\citep{zhu-2008,zalan-2011}.
Rather than clustering \bert{} embeddings, \alps{} clusters
surprisal embeddings.
We compare the clusters between surprisal embeddings and \bert{} embeddings to
understand the structure of the surprisal clusters.
First, we use t-\abr{sne}~\citep{maaten-2008} to plot the embeddings for each sentence
in the
\abr{imdb} training set (Figure~\ref{fig:embed}).
The labels are not well-separated for both embedding sets, but
the surprisal embeddings seem easier to cluster.
To quantitively measure
cluster quality, we use the Silhouette Coefficient for which larger values indicate
desirable clustering~\citep{rousseeuw-1987}. The surprisal clusters have
a coefficient of 0.38, whereas the \bert{} clusters have a coefficient of only 0.04.

These results, along with the classification experiments, show that
na\"ively clustering \bert{} embeddings is not suited for \al{}.
Possibly, more complicated clustering algorithms can capture the
intrinsic structure of the \bert{} embeddings.  However, this would increase
the algorithmic complexity and runtime.
Alternatively, one can map the feature representations to a space where simple
clustering algorithms work well.  During this transformation,
important information
for \al{} must be preserved and extracted.
Our approach uses the \mlm{} head, which has already been trained on
extensive corpora, to map the \bert{} embeddings into the surprisal
embedding space.  As a result, simple \km{} can efficiently choose representative
sentences.

\paragraph{Single-iteration Sampling}

\begin{table}[t]
    \centering
    \begin{tabular}{l rrr}
    \toprule
    Dataset & $k$ & Iterative & Single \\
    \midrule
    \abr{imdb} & 200 & 0.63 $\pm$ 0.04 &  0.61 $\pm$ 0.03 \\
    & 500 & 0.74 $\pm$ 0.05 &  0.76 $\pm$ 0.04 \\
    & 1000 & 0.82 $\pm$ 0.01 &  0.82 $\pm$ 0.01 \\
    \midrule
    \pubmed & 200 & 0.63 $\pm$ 0.03 &  0.64 $\pm$ 0.03 \\
    & 500 & 0.80 $\pm$ 0.02 &  0.82 $\pm$ 0.01 \\
    & 1000 & 0.84 $\pm$ 0.00 &  0.84 $\pm$ 0.00 \\
    \bottomrule
    \end{tabular}
    \caption{Test accuracy on \abr{imdb} and PubMed between different uses
        of \alps~for various $k$, the number of sentences to query. We compare
        using \alps~iteratively (Iterative) as done in Section~\ref{sec:experiments} with
        using \alps~to query all $k$ sentences in one iteration (Single).
  The test accuracy does not change much, showing that
\alps~is flexible to apply in different settings.}
    \label{tab:single}
\end{table}

In Section~\ref{sec:experiments}, we sample data iteratively
(Algorithm~\ref{alg:active}) to fairly compare the
different \al{} algorithms.  However, \alps~does not require
updating
the classifier because it only depends on the pre-trained encoder.  Rather than
sampling data in small batches and re-training the model, \alps~can sample a
batch of $k$ sentences in one iteration (Algorithm~\ref{alg:alps}).
Between using \alps{} iteratively and deploying the algorithm for a single iteration, the
difference is insignificant (Table~\ref{tab:single}). Plus, sampling 1{,}000
sentences only takes about 97 minutes for \pubmed{} and 7 minutes for \abr{imdb}.

With this flexibility in sampling,
\alps{} can accommodate different budget
constraints.  For example, re-training the classifier
may be costly, so users want a sampling algorithm that can query $k$ sentences all at
once.  In other cases, annotators are not always available, so the
number of obtainable annotations is unpredictable.
Then, users would prefer an \al{} strategy that can query a variable number of
sentences for any iteration.  These cases illustrate practical
needs for a cold-start algorithm like \alps{}.

\section{Related Work}
Active learning has shown success in tasks,
such as named entity recognition~\citep{shen-2004}, word sense
disambiguation~\citep{zhu-2007}, and sentiment analysis~\citep{li-2012}.
\citet{wang-2014} are the first to adapt prior \al{} work to deep learning.
However, popular heuristics~\citep{settles-2009} for querying individual
points do not work as well in a batch setting.
Since then, more research has been conducted on batch \al{} for deep learning.
\citet{zhang-2017} propose the first work on \al{} for neural text
classification.  They assume that the classifier is a convolutional neural
network and use expected gradient length~\citep{settles-2008} to choose
sentences that contain words with the most label-discriminative embeddings.
Besides text classification, \al{} has been applied to neural models for semantic
parsing~\citep{duong-2018}, named entity recognition~\citep{shen-2018}, and
machine translation~\citep{liu-2018}.

\alps~makes use of \bert, a model that excels at transfer learning.  Other
works also combine \al{} and transfer learning to select training data that
reduce generalization error.  \citet{rai-2010} measures
domain divergence from the source domain to select the most informative texts in the target domain.
\citet{xwang-2014} use \al{} to query points for a target task through matching conditional
distributions.  Additionally, combining word-level and document-level
annotations can improve knowledge transfer~\citep{settles-2011,yuan-2020-clime}.

In addition to uncertainty and diversity sampling, other areas of deep \al{} focus on
Bayesian approaches~\citep{siddhant-2018,kirsch-2019} and reinforcement
learning~\citep{fang-2017}.  An interesting research direction can
integrate one of these approaches with \alps{}.

\section{Conclusion}

Transformers are powerful models that have revolutionized \abr{nlp}.
Nevertheless,
like other deep models, their accuracy and stability require fine-tuning on
large amounts of data.
\al{} should level the playing field by directing
limited annotations most effectively so that labels complement, rather than
duplicate, unsupervised data.  Luckily, transformers have generalized knowledge
about language that can help acquire data for fine-tuning.  Like \badge{}, we project
data into an embedding space and then select the most
representative points. Our method is unique because it
only relies on self-supervision to conduct sampling.
Using the pre-trained loss guides the \al{} process to sample diverse and
uncertain examples in the cold-start setting.
  Future work may focus on finding representations that encode the most
  important information for \al{}.

\section*{Acknowledgments}

We thank Kuen-Han Tsai, Chien-Min Yu, Si-An Chen, Pedro Rodriguez, Eleftheria
Briakou, and the anonymous reviewers for their feedback.
Michelle Yuan is supported by JHU Human Language Technology Center of Excellence (HLTCOE).
Jordan Boyd-Graber is supported in part by the Office of the Director of National Intelligence (ODNI), Intelligence Advanced Research Projects Activity (IARPA), via the BETTER Program contract \#2019-19051600005. The views and conclusions contained herein are those of the authors and should not be interpreted as necessarily representing the official policies, either expressed or implied, of ODNI, IARPA, or the U.S. Government. The U.S. Government is authorized to reproduce and distribute reprints for governmental purposes notwithstanding any copyright annotation therein.

\bibliography{bib/journal-full,bib/jbg,bib/michelle}

\begin{thebibliography}{56}
\expandafter\ifx\csname natexlab\endcsname\relax\def\natexlab#1{#1}\fi

\bibitem[{Arthur and Vassilvitskii(2006)}]{arthur-2006}
David Arthur and Sergei Vassilvitskii. 2006.
\newblock k-means++: {T}he advantages of careful seeding.
\newblock Technical report, Stanford.

\bibitem[{Ash and Adams(2019)}]{ash-2019}
Jordan~T. Ash and Ryan~P. Adams. 2019.
\newblock \href {http://arxiv.org/abs/arXiv:1910.08475} {On warm-starting
  neural network training}.
\newblock \emph{arXiv preprint arXiv:1910.08475}.

\bibitem[{Ash et~al.(2020)Ash, Zhang, Krishnamurthy, Langford, and
  Agarwal}]{ash-2020}
Jordan~T. Ash, Chicheng Zhang, Akshay Krishnamurthy, John Langford, and Alekh
  Agarwal. 2020.
\newblock Deep batch active learning by diverse, uncertain gradient lower
  bounds.
\newblock In \emph{Proceedings of the International Conference on Learning
  Representations}.

\bibitem[{Aytekin et~al.(2018)Aytekin, Ni, Cricri, and Aksu}]{aytekin-2018}
Caglar Aytekin, Xingyang Ni, Francesco Cricri, and Emre Aksu. 2018.
\newblock Clustering and unsupervised anomaly detection with {L2} normalized
  deep auto-encoder representations.
\newblock In \emph{International Joint Conference on Neural Networks}.

\bibitem[{Beltagy et~al.(2019)Beltagy, Lo, and Cohan}]{beltagy-2019}
Iz~Beltagy, Kyle Lo, and Arman Cohan. 2019.
\newblock \href {https://www.aclweb.org/anthology/D19-1371.pdf} {Sci{BERT}: {A}
  pretrained language model for scientific text}.
\newblock In \emph{Proceedings of Empirical Methods in Natural Language
  Processing}.

\bibitem[{Beyer et~al.(1999)Beyer, Goldstein, Ramakrishnan, and
  Shaft}]{beyer-1999}
Kevin Beyer, Jonathan Goldstein, Raghu Ramakrishnan, and Uri Shaft. 1999.
\newblock When is ``nearest neighbor'' meaningful?
\newblock In \emph{International {C}onference on {D}atabase {T}heory}.

\bibitem[{Bod\'o et~al.(2011)Bod\'o, Minier, and Csat\'o}]{zalan-2011}
Zal\'an Bod\'o, Zsolt Minier, and Lehel Csat\'o. 2011.
\newblock Active learning with clustering.
\newblock In \emph{Active Learning and Experimental Design Workshop in
  Conjunction with AISTATS 2010}.

\bibitem[{Dasgupta(2011)}]{dasgupta-2011}
Sanjoy Dasgupta. 2011.
\newblock Two faces of active learning.
\newblock \emph{Theoretical computer science}, 412(19):1767--1781.

\bibitem[{Davison et~al.(2019)Davison, Feldman, and Rush}]{davison-2019}
Joe Davison, Joshua Feldman, and Alexander~M. Rush. 2019.
\newblock \href {https://doi.org/10.18653/v1/D19-1109} {Commonsense knowledge
  mining from pretrained models}.
\newblock In \emph{Proceedings of Empirical Methods in Natural Language
  Processing}.

\bibitem[{Dernoncourt and Lee(2017)}]{dernoncourt-2017}
Franck Dernoncourt and Ji~Young Lee. 2017.
\newblock \href {https://www.aclweb.org/anthology/I17-2052} {Pub{M}ed 200k
  {RCT}: a dataset for sequential sentence classification in medical
  abstracts}.
\newblock \emph{International Joint Conference on Natural Language Processing},
  2:308--313.

\bibitem[{Devlin et~al.(2019)Devlin, Chang, Lee, and Toutanova}]{devlin-2019}
Jacob Devlin, Ming-Wei Chang, Kenton Lee, and Kristina Toutanova. 2019.
\newblock \href {https://doi.org/10.18653/v1/N19-1423} {{BERT}: {P}re-training
  of deep bidirectional transformers for language understanding}.
\newblock In \emph{Conference of the North American Chapter of the Association
  for Computational Linguistics}.

\bibitem[{Dodge et~al.(2020)Dodge, Ilharco, Schwartz, Farhadi, Hajishirzi, and
  Smith}]{dodge-2020}
Jesse Dodge, Gabriel Ilharco, Roy Schwartz, Ali Farhadi, Hannaneh Hajishirzi,
  and Noah Smith. 2020.
\newblock \href {http://arxiv.org/abs/arXiv:2002.06305} {Fine-tuning pretrained
  language models: Weight initializations, data orders, and early stopping}.
\newblock \emph{arXiv preprint arXiv:2002.06305}.

\bibitem[{Duong et~al.(2018)Duong, Afshar, Estival, Pink, Cohen, and
  Johnson}]{duong-2018}
Long Duong, Hadi Afshar, Dominique Estival, Glen Pink, Philip~R Cohen, and Mark
  Johnson. 2018.
\newblock \href {https://doi.org/10.18653/v1/P18-2008} {Active learning for
  deep semantic parsing}.
\newblock In \emph{Proceedings of the Association for Computational
  Linguistics}.

\bibitem[{Ettinger(2020)}]{ettinger-2020}
Allyson Ettinger. 2020.
\newblock \href {https://doi.org/10.1162/tacl_a_00298} {What {BERT} is not:
  Lessons from a new suite of psycholinguistic diagnostics for language
  models}.
\newblock \emph{Transactions of the Association for Computational Linguistics},
  8:34--48.

\bibitem[{Fang et~al.(2017)Fang, Li, and Cohn}]{fang-2017}
Meng Fang, Yuan Li, and Trevor Cohn. 2017.
\newblock \href {https://doi.org/10.18653/v1/D17-1063} {Learning how to active
  learn: A deep reinforcement learning approach}.
\newblock In \emph{Proceedings of Empirical Methods in Natural Language
  Processing}.

\bibitem[{Felt et~al.(2015)Felt, Ringger, Seppi, Black, and
  Haertel}]{felt-2015}
Paul Felt, Eric Ringger, Kevin Seppi, Kevin Black, and Robbie Haertel. 2015.
\newblock \href {https://doi.org/10.3115/v1/N15-1089} {Early gains matter: {A}
  case for preferring generative over discriminativecrowdsourcing models}.
\newblock In \emph{Proceedings of the Association for Computational
  Linguistics}.

\bibitem[{Goldberg(2019)}]{goldberg-2019}
Yoav Goldberg. 2019.
\newblock \href {http://arxiv.org/abs/arXiv:1901.05287} {Assessing {BERT}'s
  syntactic abilities}.
\newblock \emph{arXiv preprint arXiv:1901.05287}.

\bibitem[{Guo et~al.(2017)Guo, Pleiss, Sun, and Weinberger}]{guo-2017}
Chuan Guo, Geoff Pleiss, Yu~Sun, and Kilian~Q. Weinberger. 2017.
\newblock On calibration of modern neural networks.
\newblock \emph{Journal of Machine Learning Research}, 70:1321--1330.

\bibitem[{Hendrycks et~al.(2020)Hendrycks, Liu, Wallace, Dziedzic, Krishnan,
  and Song}]{hendrycks-2020}
Dan Hendrycks, Xiaoyuan Liu, Eric Wallace, Adam Dziedzic, Rishabh Krishnan, and
  Dawn Song. 2020.
\newblock \href {https://doi.org/10.18653/v1/2020.acl-main.244} {Pretrained
  transformers improve out-of-distribution robustness}.
\newblock In \emph{Proceedings of the Association for Computational
  Linguistics}.

\bibitem[{Hsu and Lin(2015)}]{hsu-2015}
Wei-Ning Hsu and Hsuan-Tien Lin. 2015.
\newblock Active learning by learning.
\newblock In \emph{Association for the Advancement of Artificial Intelligence}.

\bibitem[{Hu et~al.(2010)Hu, Mac~Namee, and Delany}]{hu-2010}
Rong Hu, Brian Mac~Namee, and Sarah~Jane Delany. 2010.
\newblock Off to a good start: {U}sing clustering to select the initial
  training set in active learning.
\newblock In \emph{Florida Artificial Intelligence Research Society
  Conference}.

\bibitem[{Kirsch et~al.(2019)Kirsch, van Amersfoort, and Gal}]{kirsch-2019}
Andreas Kirsch, Joost van Amersfoort, and Yarin Gal. 2019.
\newblock {BatchBALD}: {E}fficient and diverse batch acquisition for deep
  {B}ayesian active learning.
\newblock In \emph{Proceedings of Advances in Neural Information Processing
  Systems}.

\bibitem[{Lewis and Gale(1994)}]{lewis-1994}
David~D. Lewis and William~A. Gale. 1994.
\newblock A sequential algorithm for training text classifiers.
\newblock In \emph{Proceedings of the ACM SIGIR Conference on Research and
  Development in Information Retrieval}.

\bibitem[{Li et~al.(2012)Li, Ju, Zhou, and Li}]{li-2012}
Shoushan Li, Shengfeng Ju, Guodong Zhou, and Xiaojun Li. 2012.
\newblock \href {https://www.aclweb.org/anthology/D12-1013.pdf} {Active
  learning for imbalanced sentiment classification}.
\newblock In \emph{Proceedings of Empirical Methods in Natural Language
  Processing}.

\bibitem[{Liu et~al.(2018)Liu, Buntine, and Haffari}]{liu-2018}
Ming Liu, Wray Buntine, and Gholamreza Haffari. 2018.
\newblock \href {https://doi.org/10.18653/v1/K18-1033} {Learning to actively
  learn neural machine translation}.
\newblock In \emph{Conference on Computational Natural Language Learning}.

\bibitem[{Liu et~al.(2019)Liu, Ott, Goyal, Du, Joshi, Chen, Levy, Lewis,
  Zettlemoyer, and Stoyanov}]{liu-2019}
Yinhan Liu, Myle Ott, Naman Goyal, Jingfei Du, Mandar Joshi, Danqi Chen, Omer
  Levy, Mike Lewis, Luke Zettlemoyer, and Veselin Stoyanov. 2019.
\newblock \href {http://arxiv.org/abs/arXiv:1907.11692} {{RoBERTa}: {A}
  robustly optimized bert pretraining approach}.
\newblock \emph{arXiv preprint arXiv:1907.11692}.

\bibitem[{Loshchilov and Hutter(2019)}]{loshchilov-2019}
Ilya Loshchilov and Frank Hutter. 2019.
\newblock Decoupled weight decay regularization.
\newblock In \emph{Proceedings of the International Conference on Learning
  Representations}.

\bibitem[{Lowell et~al.(2019)Lowell, Lipton, and Wallace}]{lowell-2019}
David Lowell, Zachary~C. Lipton, and Byron~C. Wallace. 2019.
\newblock \href {https://doi.org/10.18653/v1/D19-1003} {Practical obstacles to
  deploying active learning}.
\newblock In \emph{Proceedings of Empirical Methods in Natural Language
  Processing}.

\bibitem[{Maas et~al.(2011)Maas, Daly, Pham, Huang, Ng, and Potts}]{maas-2011}
Andrew~L. Maas, Raymond~E. Daly, Peter~T. Pham, Dan Huang, Andrew~Y. Ng, and
  Christopher Potts. 2011.
\newblock \href {https://www.aclweb.org/anthology/P11-1015} {Learning word
  vectors for sentiment analysis}.
\newblock In \emph{Proceedings of the Association for Computational
  Linguistics}.

\bibitem[{Maaten and Hinton(2008)}]{maaten-2008}
Laurens van~der Maaten and Geoffrey Hinton. 2008.
\newblock Visualizing data using t-{SNE}.
\newblock \emph{Journal of Machine Learning Research}, 9:2579--2605.

\bibitem[{Petroni et~al.(2019)Petroni, Rockt{\"a}schel, Riedel, Lewis, Bakhtin,
  Wu, and Miller}]{petroni-2019}
Fabio Petroni, Tim Rockt{\"a}schel, Sebastian Riedel, Patrick Lewis, Anton
  Bakhtin, Yuxiang Wu, and Alexander Miller. 2019.
\newblock \href {https://www.aclweb.org/anthology/D19-1250.pdf} {Language
  models as knowledge bases?}
\newblock In \emph{Proceedings of Empirical Methods in Natural Language
  Processing}.

\bibitem[{Rai et~al.(2010)Rai, Saha, Daum{\'e}~III, and
  Venkatasubramanian}]{rai-2010}
Piyush Rai, Avishek Saha, Hal Daum{\'e}~III, and Suresh Venkatasubramanian.
  2010.
\newblock \href {https://www.aclweb.org/anthology/W10-0104} {Domain adaptation
  meets active learning}.
\newblock In \emph{Conference of the North American Chapter of the Association
  for Computational Linguistics}.

\bibitem[{Rousseeuw(1987)}]{rousseeuw-1987}
Peter~J. Rousseeuw. 1987.
\newblock Silhouettes: A graphical aid to the interpretation and validation of
  cluster analysis.
\newblock \emph{Journal of Computational and Applied Mathematics}, 20:53--65.

\bibitem[{Salazar et~al.(2020)Salazar, Liang, Nguyen, and
  Kirchhoff}]{salazar-2020}
Julian Salazar, Davis Liang, Toan~Q. Nguyen, and Katrin Kirchhoff. 2020.
\newblock \href {https://doi.org/10.18653/v1/2020.acl-main.240} {Masked
  language model scoring}.
\newblock In \emph{Proceedings of the Association for Computational
  Linguistics}.

\bibitem[{Sener and Savarese(2018)}]{sener-2018}
Ozan Sener and Silvio Savarese. 2018.
\newblock Active learning for convolutional neural networks: {A} core-set
  approach.
\newblock In \emph{Proceedings of the International Conference on Learning
  Representations}.

\bibitem[{Settles(2009)}]{settles-2009}
Burr Settles. 2009.
\newblock Active learning literature survey.
\newblock Technical report, University of Wisconsin-Madison Department of
  Computer Sciences.

\bibitem[{Settles(2011)}]{settles-2011}
Burr Settles. 2011.
\newblock \href {https://www.aclweb.org/anthology/D11-1136} {Closing the loop:
  Fast, interactive semi-supervised annotation with queries on features and
  instances}.
\newblock In \emph{Proceedings of Empirical Methods in Natural Language
  Processing}.

\bibitem[{Settles et~al.(2008)Settles, Craven, and Ray}]{settles-2008}
Burr Settles, Mark Craven, and Soumya Ray. 2008.
\newblock Multiple-instance active learning.
\newblock In \emph{Proceedings of Advances in Neural Information Processing
  Systems}.

\bibitem[{Shannon(1948)}]{shannon-1948}
Claude~Elwood Shannon. 1948.
\newblock A mathematical theory of communication.
\newblock \emph{Bell system technical journal}, 27.

\bibitem[{Shen et~al.(2004)Shen, Zhang, Su, Zhou, and Tan}]{shen-2004}
Dan Shen, Jie Zhang, Jian Su, Guodong Zhou, and Chew-Lim Tan. 2004.
\newblock \href {https://www.aclweb.org/anthology/P04-1075.pdf}
  {Multi-criteria-based active learning for named entity recognition}.
\newblock In \emph{Proceedings of the Association for Computational
  Linguistics}.

\bibitem[{Shen et~al.(2018)Shen, Yun, Lipton, Kronrod, and
  Anandkumar}]{shen-2018}
Yanyao Shen, Hyokun Yun, Zachary~C. Lipton, Yakov Kronrod, and Animashree
  Anandkumar. 2018.
\newblock Deep active learning for named entity recognition.
\newblock In \emph{Proceedings of the International Conference on Learning
  Representations}.

\bibitem[{Siddhant and Lipton(2018)}]{siddhant-2018}
Aditya Siddhant and Zachary~C. Lipton. 2018.
\newblock \href {https://doi.org/10.18653/v1/D18-1318} {Deep bayesian active
  learning for natural language processing: {R}esults of a large-scale
  empirical study}.
\newblock In \emph{Proceedings of Empirical Methods in Natural Language
  Processing}.

\bibitem[{Socher et~al.(2013)Socher, Perelygin, Wu, Chuang, Manning, Ng, and
  Potts}]{socher-2013}
Richard Socher, Alex Perelygin, Jean Wu, Jason Chuang, Christopher~D. Manning,
  Andrew~Y. Ng, and Christopher Potts. 2013.
\newblock \href {https://www.aclweb.org/anthology/D13-1170} {Recursive deep
  models for semantic compositionality over a sentiment treebank}.
\newblock In \emph{Proceedings of Empirical Methods in Natural Language
  Processing}.

\bibitem[{Strubell et~al.(2019)Strubell, Ganesh, and McCallum}]{strubell-2019}
Emma Strubell, Ananya Ganesh, and Andrew McCallum. 2019.
\newblock \href {https://doi.org/10.18653/v1/P19-1355} {Energy and policy
  considerations for deep learning in {NLP}}.
\newblock In \emph{Proceedings of the Association for Computational
  Linguistics}.

\bibitem[{Tenney et~al.(2019)Tenney, Xia, Chen, Wang, Poliak, McCoy, Kim,
  Van~Durme, Bowman, Das et~al.}]{tenney-2019}
Ian Tenney, Patrick Xia, Berlin Chen, Alex Wang, Adam Poliak, R.~Thomas McCoy,
  Najoung Kim, Benjamin Van~Durme, Samuel~R. Bowman, Dipanjan Das, et~al. 2019.
\newblock What do you learn from context? {P}robing for sentence structure in
  contextualized word representations.
\newblock In \emph{Proceedings of the International Conference on Learning
  Representations}.

\bibitem[{Tomasev et~al.(2013)Tomasev, Radovanovic, Mladenic, and
  Ivanovic}]{tomasev-2013}
Nenad Tomasev, Milos Radovanovic, Dunja Mladenic, and Mirjana Ivanovic. 2013.
\newblock The role of hubness in clustering high-dimensional data.
\newblock \emph{IEEE {T}ransactions on {K}nowledge and {D}ata {E}ngineering},
  26(3):739--751.

\bibitem[{Voorhees et~al.(2020)Voorhees, Alam, Bedrick, Demner-Fushman, Hersh,
  Lo, Roberts, Soboroff, and Wang}]{voorhees-2020}
Ellen Voorhees, Tasmeer Alam, Steven Bedrick, Dina Demner-Fushman, William~R.
  Hersh, Kyle Lo, Kirk Roberts, Ian Soboroff, and Lucy~Lu Wang. 2020.
\newblock \href {http://arxiv.org/abs/arXiv:2005.04474} {{TREC-COVID}:
  Constructing a pandemic information retrieval test collection}.
\newblock \emph{arXiv preprint arXiv:2005.04474}.

\bibitem[{Wang and Shang(2014)}]{wang-2014}
Dan Wang and Yi~Shang. 2014.
\newblock A new active labeling method for deep learning.
\newblock In \emph{International Joint Conference on Neural Networks}.

\bibitem[{Wang et~al.(2014)Wang, Huang, and Schneider}]{xwang-2014}
Xuezhi Wang, Tzu-Kuo Huang, and Jeff Schneider. 2014.
\newblock Active transfer learning under model shift.
\newblock In \emph{Proceedings of the International Conference on Learning
  Representations}.

\bibitem[{Xu et~al.(2003)Xu, Yu, Tresp, Xu, and Wang}]{xu-2003}
Zhao Xu, Kai Yu, Volker Tresp, Xiaowei Xu, and Jizhi Wang. 2003.
\newblock Representative sampling for text classification using support vector
  machines.
\newblock In \emph{Proceedings of the European Conference on Information
  Retrieval}.

\bibitem[{Yang et~al.(2019)Yang, Dai, Yang, Carbonell, Salakhutdinov, and
  Le}]{yang-2019}
Zhilin Yang, Zihang Dai, Yiming Yang, Jaime Carbonell, Ruslan Salakhutdinov,
  and Quoc~V. Le. 2019.
\newblock {XLNet}: Generalized autoregressive pretraining for language
  understanding.
\newblock In \emph{Proceedings of Advances in Neural Information Processing
  Systems}.

\bibitem[{Yuan et~al.(2020)Yuan, Zhang, Durme, Findlater, and
  Boyd-Graber}]{yuan-2020-clime}
Michelle Yuan, Mozhi Zhang, Benjamin~Van Durme, Leah Findlater, and Jordan
  Boyd-Graber. 2020.
\newblock Interactive refinement of cross-lingual word embeddings.
\newblock In \emph{Proceedings of Empirical Methods in Natural Language
  Processing}.

\bibitem[{Zhang et~al.(2015)Zhang, Zhao, and LeCun}]{zhang-2015}
Xiang Zhang, Junbo Zhao, and Yann LeCun. 2015.
\newblock Character-level convolutional networks for text classification.
\newblock In \emph{Proceedings of Advances in Neural Information Processing
  Systems}.

\bibitem[{Zhang et~al.(2017)Zhang, Lease, and Wallace}]{zhang-2017}
Ye~Zhang, Matthew Lease, and Byron~C. Wallace. 2017.
\newblock Active discriminative text representation learning.
\newblock In \emph{Association for the Advancement of Artificial Intelligence}.

\bibitem[{Zhu and Hovy(2007)}]{zhu-2007}
Jingbo Zhu and Eduard Hovy. 2007.
\newblock \href {https://www.aclweb.org/anthology/D07-1082} {Active learning
  for word sense disambiguation with methods for addressing the class imbalance
  problem}.
\newblock In \emph{Proceedings of Empirical Methods in Natural Language
  Processing}.

\bibitem[{Zhu et~al.(2008)Zhu, Wang, Yao, and Tsou}]{zhu-2008}
Jingbo Zhu, Huizhen Wang, Tianshun Yao, and Benjamin~K. Tsou. 2008.
\newblock \href {https://www.aclweb.org/anthology/C08-1143} {Active learning
  with sampling by uncertainty and density for word sense disambiguation and
  text classification}.
\newblock In \emph{Proceedings of International Conference on Computational
  Linguistics}.

\end{thebibliography}
\bibliographystyle{style/acl_natbib}

\clearpage
\appendix
\section{Appendices}
\label{sec:appendix}

\begin{table*}[h]
    \centering
    \begin{tabular}{l rr rr}
    \toprule
    & \multicolumn{2}{c}{IMDB} &
    \multicolumn{2}{c}{SST-2} \\
    \cmidrule(lr){2-3} \cmidrule(lr){4-5}
    & $k=100$ & $k=200$ & $k=100$ & $k=200$\\
    \midrule
        \alps & $0.60 \pm 0.03$& \bm{$0.69 \pm 0.04$} & \bm{$0.57 \pm
        0.06$} & \bm{$0.64 \pm 0.04$}  \\
        \alps-tokens-0.1 & \bm{$0.61 \pm 0.05$} & $0.63 \pm 0.11$ &
        $0.56 \pm 0.07$ & $0.63 \pm 0.04$ \\
        \alps-tokens-0.2 & $0.55 \pm 0.07$ & $0.65 \pm 0.05$ & \bm{$0.57 \pm
        0.05$} & $0.63 \pm
        0.05$ \\
        \alps-tokens-1.0 & $0.59 \pm 0.05$ & $0.65 \pm 0.07$ & $0.56 \pm
        0.05$ & $0.62 \pm 0.05$\\
        \alps-masked & $0.59 \pm 0.03$ & $0.63 \pm 0.09$ & $0.56 \pm
        0.03$ & $0.60 \pm
        0.02$\\
    \bottomrule
    \end{tabular}
    \caption{Comparison of validation accuracy between the variants of \alps~to sample data for
        \abr{imdb} and SST-2  in the first two iterations. \alps-tokens-$p$ varies the percentage
    $p$ of
    tokens evaluated with \mlm{} loss when computing surprisal embeddings.  \alps-masked
    passes in the input with masks as originally done in pre-training.
    Overall, we observe that \alps{} has higher mean and smaller variance in
    accuracy.}
    \label{tab:mask_acc}
\end{table*}

\begin{figure}[h]
    \centering
    \includegraphics[width=\linewidth]{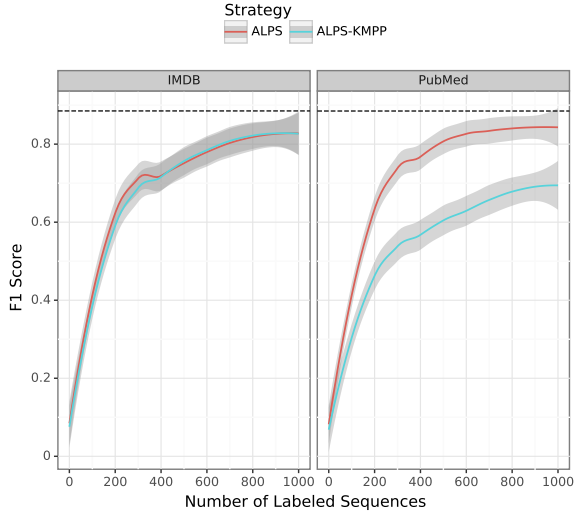}
    \caption{Comparing validation accuracy between using \km{} and \kmpp{}
    to select centroids in the surprisal embeddings.  Using \km{}
reaches higher accuracy.}
    \label{fig:km_kp}
\end{figure}

\begin{figure}[t]
\centering
    \begin{subfigure}{\linewidth}
        \centering
    \includegraphics[width=0.8\linewidth]{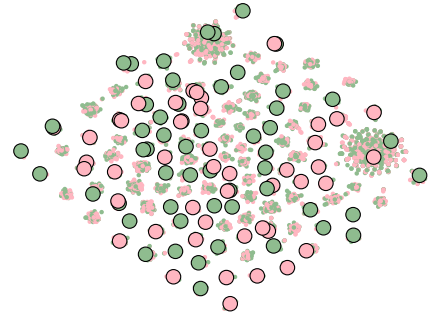}
    \caption{Surprisal embeddings with \kmpp~centers}
    \label{fig:mlmkp}
    \end{subfigure}
    \begin{subfigure}{\linewidth}
        \centering
    \includegraphics[width=0.8\linewidth]{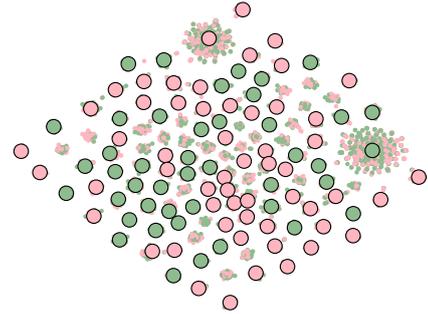}
    \caption{Surprisal embeddings with \km~centers}
    \label{fig:mlmkm}
    \end{subfigure}
    \caption{\abr{t-sne} plots of
        surprisal embeddings for \abr{imdb} training data.  The centers are either picked by \kmpp{} (right) or \km{} (left).
     There is less overlap between the centers with \km{}
    compared to \kmpp{}.  So, using \km{} is better for exploiting
    diversity in the surprisal embedding space.
}
\label{fig:embedkmpp}
\end{figure}

\subsection{Token Masking}
\label{ssec:mask}
In our preliminary experiments on the validation set, we notice improvement in
accuracy after passing in
the original input with no masks (Table~\ref{tab:mask_acc}).  The purpose of the \texttt{[MASK]} token during
pre-training is to train the token embeddings to learn context so that it can
predict the token labels.  Since we are not training the token embeddings to
learn context, masking the tokens does not help much for \al{}.
We use \al{} for fine-tuning, so the input should be in the same format for
\al{} and fine-tuning.  Otherwise, there is a mismatch between the two stages.

\subsection{Token Sampling for Evaluation}
\label{ssec:sample}
When \bert~evaluates MLM loss, it only focuses on the masked tokens,
which are from a 15\% random subsample of tokens in the sentence.  We experiment
with varying this subsample percentage on the validation set
(Table~\ref{tab:mask_acc}).  We try sampling 10\%, 15\%, 20\%, and 100\%.
Overall, we notice that mean accuracy are roughly the same, but variance in
accuracy across different runs is slightly higher for percentages other than
15\%.

After the
second \al{} iteration, we notice that accuracy mean and variance between the
different token sampling percentages converge.  So, the token sampling
percentage makes more of a difference in early stages of \al.
\citet{devlin-2019} show that the difference in accuracy between various mask strategies is
minimal for fine-tuning \bert.  We believe this can also be applied to what we
have observed for \alps.

\subsection{\km{} vs. \kmpp{}}
\label{ssec:km}
The state-of-the-art baseline \badge{} applies \kmpp{} on gradient
embeddings to select points to query.  Initially, we also use \kmpp{} on the
surprisal embeddings but validation accuracy is only slightly higher
than random sampling.  Since \kmpp{} is originally an algorithm for robust
initialization of \km{}, we instead apply \km{} on the surprisal embeddings.  As
a result, we see more significant increase in accuracy over baselines,
especially for PubMed (Figure~\ref{fig:km_kp}).  Additionally, the t-\abr{sne} plots
show that \km{} selects centers that are further apart compared to the
ones chosen by \kmpp{} (Figure~\ref{fig:embedkmpp}).  This shows that \km{} can help sample a more diverse
batch of data.

\begin{table*}[t]
    \centering
    \begin{tabularx}{\textwidth}{lXX}
    \toprule
    & \agnews & \pubmed \\
    \midrule
    \multirow{2}{*}{\alps} & Jason Thomas matches a career-high with \hl{26}
    points \hl{and} \hl{American} \hl{wins} its
    fifth straight \hl{by} \hl{beating} \hl{visiting} Ohio, 64-55, Saturday at Bender Arena
    (Sports)&
    The results showed that physical activity and exercise capacity in the
    \hl{intervention} group was significantly \hl{higher} \hl{than} the control group after the
    intervention . (results)\\
    & Sainsbury says it will \hl{take} \hl{a} 550 \hl{million} pound hit to profits
    \hl{this} year
    as it invests to boost sales and \hl{reverse} falling \hl{market} share (Business) &
    Flu\hl{maz}eni\hl{l} was \hl{administered} after the \hl{completion} of endoscopy under sedation to reduce recovery time and increase patient safety .
    (objective) \\
    \multirow{2}{*}{Random} & & \\
    & Bernhard Langer and Hal Sutton stressed the importance of playing this
    year's 135th Ryder Cup \dots
    (Sports)
    & The study population consisted of 20 interns and medical students
    (methods) \\
    & BLOOMFIELD TOWNSHIP, Mich. -- When yesterday's Ryder Cup pairings were
    announced, Bernhard Langer knew his team had been given an opportunity.
    (Sports)
    & The subject , health care provider , and research staff were blinded to
    the treatment . (methods) \\
    \bottomrule
    \end{tabularx}
    \caption{Sample sentences from AG News and PubMed while using \alps~and
    Random in the first iteration.  For \alps, highlighted tokens are the ones
that have a nonzero entry in the surprisal embedding.  Compared to random sampling,
\alps~samples sentences with more diverse content.}
    \label{tab:samples}
\end{table*}

\subsection{Sample Sentences}

Section~\ref{sec:analysis} quantitatively analyzes diversity of
\alps{}.  Here, we take a closer look at the kind of sentences that are sampled
by \alps{}.
Table~\ref{tab:samples} compares sentences that are chosen by \alps{} and random
sampling
in the first \abr{al} iteration.  The tokens highlighted are the ones
evaluated with surprisal loss.
Random sampling can fall prey to data idiosyncracies. For example, AG News has
sixty-two articles about the German golfer Bernhard Langer, and random sampling
picks multiple articles about him on one of five runs.
For PubMed, many sentences labeled as ``methods'' are simple sentences
with a short, independent clause.  While random sampling chooses many
sentences of this form, \alps~seems to avoid this problem.
Since the surprisal embedding encodes the fluctuation in information
content across the sentence, \alps{} is less likely to repeatedly choose
sentences with similar patterns in surprisal.  This may possibly diversify
syntactic structure in a sampled batch.

\end{document}